\documentclass[11pt]{article}

\usepackage[]{latex/acl}

\usepackage{times}
\usepackage{latexsym}

\usepackage[T1]{fontenc}
\usepackage[utf8]{inputenc}

\usepackage{microtype}

\usepackage{inconsolata}
\usepackage{amsmath}
\usepackage{booktabs}
\usepackage{tabularx}
\usepackage{caption}
\usepackage{graphicx}
\usepackage{epstopdf}
\usepackage{array}
\usepackage{subcaption}
\usepackage{xcolor}
\usepackage{multirow}
\usepackage{makecell}
\usepackage{float}
\usepackage[bottom]{footmisc}

\usepackage{pifont} % 提供特殊符号命令
\usepackage{xcolor} % 提供颜色选项

\usepackage{times}
\usepackage{helvet}
\usepackage{courier}
\usepackage{url}
\usepackage{amssymb}
\usepackage{tabularx}
\usepackage{enumitem}

\title{Augmenting Black-box LLMs with Medical Textbooks for Biomedical Question Answering\thanks{Published in Findings of EMNLP 2024}}

\author{
Yubo Wang, Xueguang Ma, Wenhu Chen\thanks{Corresponding author} \\
University of Waterloo \\
\texttt{\{y726wang, x93ma, wenhuchen\}@uwaterloo.ca}
}
\begin{document}
\maketitle

\begin{abstract}

Large Language Models (LLMs) like ChatGPT have demonstrated impressive abilities in generating responses based on human instructions.
However, their use in the medical field can be challenging due to their lack of specific, in-depth knowledge.
In this study, we present a system called LLMs Augmented with Medical Textbooks (LLM-AMT) designed to enhance the proficiency of LLMs in specialized domains.
LLM-AMT integrates authoritative medical textbooks into the LLMs' framework using plug-and-play modules.
These modules include a \textit{Query Augmenter}, a \textit{Hybrid Textbook Retriever}, and a \textit{Knowledge Self-Refiner}.
Together, they incorporate authoritative medical knowledge. Additionally, an \textit{LLM Reader} aids in contextual understanding.
Our experimental results on three medical QA tasks demonstrate that LLM-AMT significantly improves response quality, with accuracy gains ranging from 11.6\% to 16.6\%.
Notably, with GPT-4-Turbo as the base model, LLM-AMT outperforms the specialized Med-PaLM 2 model pre-trained on a massive amount of medical corpus by 2-3\%.
We found that despite being 100$\times$ smaller in size, medical textbooks as a retrieval corpus are proven to be a more effective knowledge database than Wikipedia in the medical domain, boosting performance by 7.8\%-13.7\%. 
The code is available online\footnote{\url{https://github.com/TIGER-AI-Lab/LLM-AMT}}.

\end{abstract}

\section{Introduction}
Recent advancements in Large Language Models (LLMs) have opened new possibilities in AI for the medical domain, enabling them to comprehend and communicate through language. The promise of these models is underscored by their performance on medical question-answering datasets \cite{zhang2018medical, pal2022medmcqa, jin2019pubmedqa}. 

LLMs are typically trained to encode world knowledge in their parameters. However, this can lead to information loss and "memory distortion" \cite{peng2023check}, resulting in the generation of plausible but incorrect content. Augmenting LLMs with external knowledge has become an interest to mitigate this, but fine-tuning LLM parameters for this purpose is often costly, especially as model sizes increase \cite{luo2022biogpt, gao2022neural, singhal2023towards}.

\begin{figure*}[ht]
\centering
\includegraphics[scale=0.19]{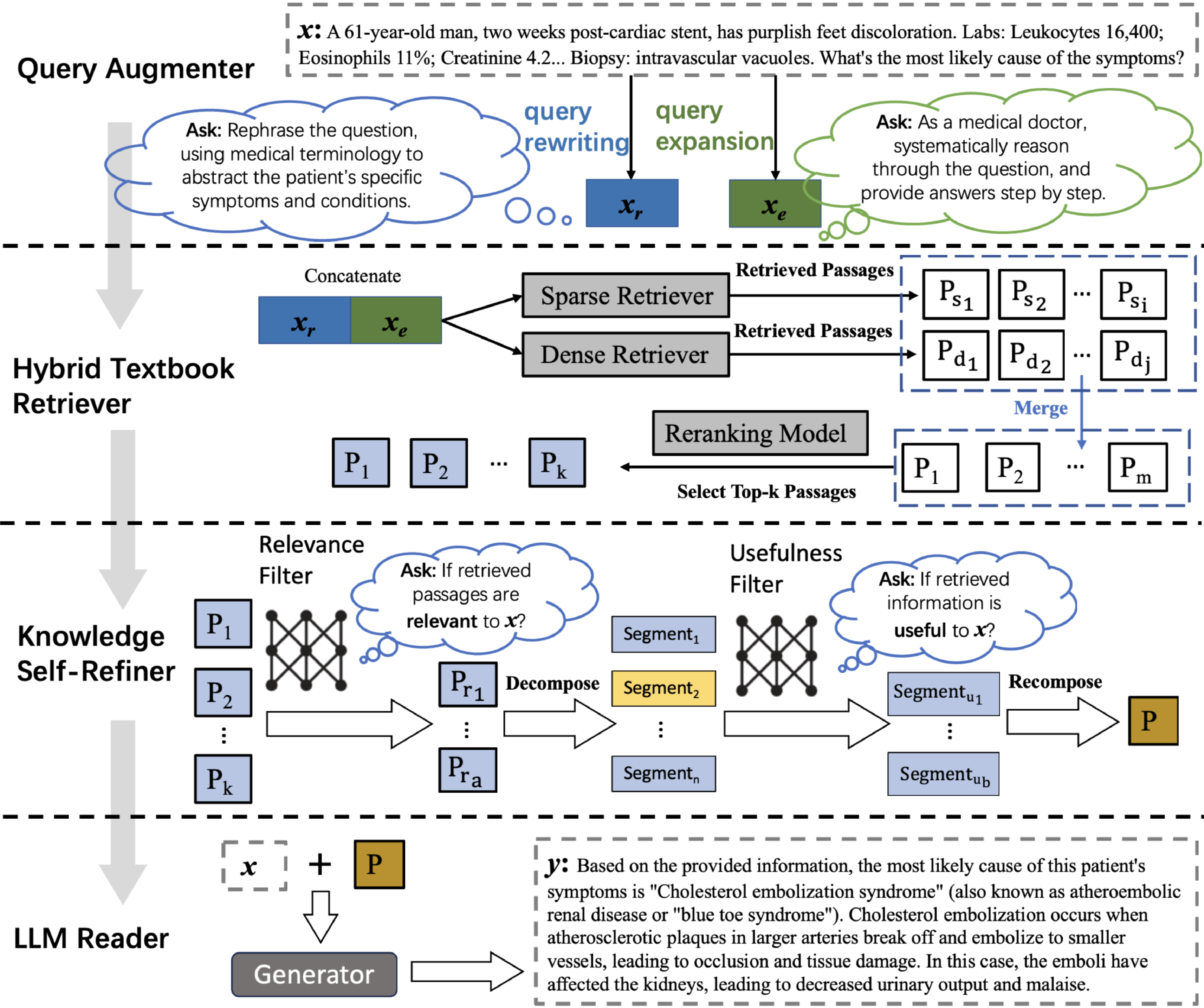}
\caption{Overview of our proposed pipeline. GPT-3.5-Turbo is prompted to perform a series of tasks including query rewriting, query expansion, relevance evaluation, and usefulness evaluation. Details of the case presented in the figure can be found in Appendix~\ref{appendix: case study}.}
\label{overview}
\end{figure*}

The Retrieval-Augmented Generation (RAG) framework provides an efficient solution to the limitations of fine-tuning in open-domain QA, pairing a retriever for sourcing relevant documents with a reader for answer extraction \cite{lewis2020retrieval, karpukhin2020dense, izacard2022few}. Enhancements in retrieval accuracy \cite{wu2021sentence, izacard2021unsupervised} and reader model co-training \cite{lewis2020retrieval, izacard2022few} have been made, with current iterations leveraging LLMs as readers to adapt specifically to their capabilities \cite{shi2023replug}. However, many rely on general knowledge bases like Wikipedia or search engines such as Google and Bing. Such sources, while vast, might lack depth in domain-specific areas like medical or financial fields. Tapping into specialized resources, such as authoritative textbooks, could yield deeper insights into complex domains.

The effectiveness of the retrieval process in enhancing LLMs with additional information is heavily reliant on the quality of retrieval.
If the retrieval process is inaccurate or contains misinformation, the utility of the RAG process can be significantly influenced~\cite{li2022survey, tan2022tegtok, shi2023large}.
To address these challenges, several approaches have been proposed, such as HyDE~\cite{gao2022precise} and query2doc~\cite{wang2023query2doc}, which aim to improve retrieval by generating hypothetical documents to expand the query.
On the other hand, methods like self-RAG~\cite{asai2023self} have introduced retrieval results reflection to filter the retrieved information for better generation.
Building on these advancements, we have developed Query Augmenter and Knowledge Self-Refiner specifically tailored for the medical RAG paradigm. 
Combining with Hybrid Textbook Retriever, we have created a novel framework, LLM-AMT, specifically designed for the medical domain. This framework leverages medical textbooks to provide LLMs with high-quality, reliable knowledge. This approach ensures that the information used during the generation process is not only accurate but also highly relevant to answer the questions.

Our evaluations across MedQA-USMLE, MedQA-MCMLE, and MedMCQA datasets demonstrate LLM-AMT's outperformance over GPT-3.5, achieving 11.6\% to 16.6\% higher accuracy. Notably, our approach leverages the in-depth knowledge from textbooks, eclipsing Wikipedia's broader scope and a 7.8\% to 13.7\% accuracy gain. With GPT-4-Turbo as a base, LLM-AMT further exceeds the medically pre-trained Med-PaLM 2 by 2.3\% to 2.7\%. Additionally, human evaluations reveal a significant 16\% reduction in hallucination occurrences during open-ended QA tasks, showcasing the model's improved reliability.

% Our contributions are fourfold:
% \begin{enumerate}[leftmargin=*,itemsep=0ex]
%     \item We propose LLM-AMT, an LLM pipeline augmented with medical textbooks, which enhances model accuracy and domain-specific expertise.
%     \item We introduce the Knowledge Self-Refiner, a novel component that implements self-refinement mechanisms for RAG models within the medical domain.
%     \item We empirically demonstrate the significant impact of domain-specific textbooks on LLM performance through comprehensive experiments, opening new avenues for research in specialized knowledge integration.
%     \item Our ablation study systematically analyzes the pipeline's architecture, evaluating the critical roles of knowledge retrieval, query augmentation, and self-refinement components in a medical context.
% \end{enumerate}
Our contributions are fourfold:
\begin{enumerate}[leftmargin=*,itemsep=0ex]
    \item We propose LLM-AMT, an LLM pipeline augmented with medical textbooks, which enhances model accuracy and domain-specific expertise.
    \item We introduce the Knowledge Self-Refiner, a novel component that implements self-refinement mechanisms for RAG models within the medical domain.
    \item We demonstrate the significant impact of domain-specific textbooks on LLM performance through comprehensive experiments, opening new avenues for research in specialized knowledge integration.
    \item We conduct an ablation study analyzing the roles of key components in our medical pipeline.
\end{enumerate}

\section{LLM-AMT}
In this paper, we introduce LLM-AMT, a dedicated process tailored for answering biomedical questions. Figure~\ref{overview} provides an overview of the process pipeline, which consists of four main components. The \textbf{Query Augmenter} rewrites and expands the input question $\boldsymbol{x}$ into a rewritten version $\boldsymbol{x}_r$ and an expanded version $\boldsymbol{x}_e$. Following this, the \textbf{Textbook Retriever} collects related passages $P_1, P_2, \ldots, P_k$ from textbooks by concatenating the augmented queries $\boldsymbol{x}_r$ and $\boldsymbol{x}_e$. Next, the \textbf{Knowledge Self-Refiner} employs a relevance filter to remove non-pertinent passages and then applies a usefulness filter to discard unhelpful segments from the remaining passages, producing refined knowledge $P$. Finally, the \textbf{LLM Reader} utilizes this refined knowledge to construct the final answer.

\subsection{Query Augmenter}

To address the challenges in biomedical question answering, where non-standard terms and discrete numerical values often impede effective information retrieval, we introduce a query augmenting module tailored for the medical domain. Our augmenter enhances queries by transforming ambiguous language and integrating key medical terms, which are crucial for accurate retrieval.

The module consists of two principal components: \textbf{query rewriting} and \textbf{query expansion}. The motivation behind query rewriting is to map colloquial or non-standard expressions to standardized medical terminology. For instance, our system converts phrases such as “high blood cell count” to the precise term “leukocytosis.” Such transformation is pivotal because it aligns patient-described symptoms with professional language. This method ensures the retention of crucial information and translates it into the language of medicine, thus making the query more suitable for professional databases.

On the other hand, query expansion leverages the LLM's ability to reason through problems without external evidence, invoking a chain-of-thought approach. By instructing the LLM with \textit{``As a medical doctor, systematically reason through the question, and provide answers step by step.''} We introduce additional relevant medical terms into the query. This preemptive reasoning extracts more directions for retrieval and enhances the likelihood of accessing pertinent information.

% Ultimately, we concatenate the rewritten and expanded parts to form the augmented query \(q\). This Query Augmenter not only captures a broader spectrum of relevant information but also significantly refines the retrieval process. By aligning patient language with medical terminology and preemptively reasoning through potential answers, our approach ensures nuanced retrieval essential for clinical question challenges.

\subsection{Textbook Retrieval Corpus}
Medical textbooks, as the epitome of knowledge in human medicine, serve as an invaluable external knowledge source. While knowledge bases like Wikipedia provide general information, textbooks offer richer and more specialized domain knowledge. In contrast to search engines like Google Search or Bing Search, the information in textbooks is more reliable. Furthermore, textbooks offer clear and concise information, making them a reliable source for text-based retrieval. For our study, we eliminated irrelevant information, such as diagrams and references, to ensure a focused, text-centric corpus. Additionally, longer paragraphs in the textbook were broken down according to periods to obtain the smallest unit for retrieval, making it easier for the LLM reader to use them as context for questions. In this paper, we utilized 51 textbooks from the MedQA dataset \cite{jin2021disease}, which are designated as the official preparation materials for the medical licensing exams. 

An overview of the statistics for the document collection in both the textbooks and Wikipedia can be seen in Table~\ref{tab: textbook and wiki comparison}. Our textbook corpus is substantially smaller in scale than Wikipedia. While Wikipedia comprises millions of paragraphs and billions of tokens, the textbook corpus, though specialized, contains fewer than 350,000 paragraphs and just over 27 million tokens. This size difference emphasizes the textbooks' concentrated domain-specific knowledge.

\begin{table}[t]
\centering
\scalebox{0.85}{
\begin{tabular}{lrr}
\toprule
\textbf{Metric} & \textbf{Textbooks} & \textbf{Wikipedia}\\
\midrule
\# of paragraphs & 347,797 & 21,015,324 \\
\# of tokens & 27,458,075 & 2,162,169,361 \\
\bottomrule
\end{tabular}}
\caption{Overall statistics of the document collection in textbooks and Wikipedia. The Wikipedia dump is from the DPR work~\cite{karpukhin2020dense}, where Wikipedia documents are split into 100-word units.}
\label{tab: textbook and wiki comparison}
\end{table}

\subsection{HybTextR (Hybrid Textbook Retriever)}
We integrate various types of retrievers in our textbook retrieval module to optimize performance, which we refer to as the HybTextR. 
For \textbf{sparse retrieval}, we follow the SPLADE \cite{formal2021splade} method. The query and document are encoded separately by BERT, and the MLM layer representation (with dimension 30k) for each token is aggregated at the maximum as the text representation. The ReLU function is used to truncate the weights in the representation to be non-negative so that it can fit into an inverted index after quantization at search time. The sparsity of this representation is effectively managed by a FLOP loss during the training stage.
For \textbf{dense retrieval}, we follow the standard pipeline proposed in the DPR work, where the query and document embeddings are taken from the CLS token's dense representation in the last layer output (with dimension 768). At search time, a k-NN search is conducted to retrieve the top relevant passages for the given query. For \textbf{reranking}, our model is a BERT-based cross-encoder. In line with the approach described in Tevatron by Gao et al. \cite{Gao2022TevatronAE}, we concatenate the query and the retrieved passages using the [SEP] token and employ the representation from the [CLS] token to predict relevance scores. This cross-encoder setup allows for a more nuanced understanding of the relationship between the query and the passages, leading to improved ranking accuracy.

% For late interaction retrieval, we use the ColBERT setting. For each query token, the token to document similarity is the maximum of the dot products of the query token to all tokens in documents. The query--document similarity score is aggregated by summation over the similarity score of all query tokens.

A core problem in the task is how to create supervised data for the neural retriever. As there is no human relevance judgment for the passages, we treat ``helpful'' passages as positive passages.
We first identify questions that GPT-3.5-Turbo answers incorrectly when provided without any contextual evidence. Then, using BM25, we recall \( n \) passages, where \( n=32 \), and concatenate each of them with the original question to serve as its context. 
Subsequently, GPT-3.5-Turbo is prompted to answer this question. Passages resulting in correct answers are treated as positive samples, whereas those leading to incorrect answers are categorized as hard negatives. Additionally, a subset of passages is randomly chosen to act as easy negative samples.

In the full pipeline of our knowledge retrieval stage, we utilize a fusion of sparse retrieval and dense retrieval as the first-phase recall model. Specifically, we merge and deduplicate passages returned by the sparse retriever (from $P_{s_1}$ to $P_{s_i}$) and those returned by the dense retriever (from $P_{d_1}$ to $P_{d_j}$), resulting in a total of $m$ unique passages as illustrated by Figure~\ref{overview}. These passages are then reordered by the reranker. Finally, the top-$k$ passages are selected for further processing. In this study, $i$, $j$, and $k$ are all set to a fixed value of $32$.

\subsection{Knowledge Self-Refiner}
The motivation for introducing the Knowledge Self-Refiner is driven by the structure of medical textbooks, which typically present elongated passages dense with information on a particular topic. Given that not all content within these passages is pertinent to addressing specific questions, and unfiltered content may lead to a diffusion of the LLM Reader's focus, we implemented the Knowledge Self-Refiner to streamline the information.

Given the retrieved passages \( P_1, P_2, \ldots, P_k \), our Knowledge Self-Refiner begins by applying a relevance filter to exclude off-topic passages. This filter performs a binary classification on each passage to determine its relevance, resulting in a subset \( P_{r_1}, P_{r_2}, \ldots, P_{r_a} \). Subsequent to this initial filtering, a decompose-then-recompose algorithm is employed. Passages are segmented at sentence boundaries according to heuristic rules. Segments that do not meet a minimum length threshold are merged with adjacent segments to ensure the combined length does not exceed 80 words. This process yields segments \( segment_1, segment_2, \ldots, segment_n \), each balancing substance with brevity.

These segments are then passed through a stringent usefulness filter, which performs a binary classification to assess the utility of each segment. This filter distills the content to retain only the segments \( segment_{u_1}, segment_{u_2}, \ldots, segment_{u_b} \) that are deemed most useful for the LLM to generate accurate responses. The final refined knowledge set, denoted by \(\textbf{P}\), is composed of these selected segments, offering a concentrated and relevant reservoir of information for the LLM.

This two-tiered filtering approach is specifically designed to address the high density of knowledge points in medical texts, which typically feature lengthy passages with only a few sentences of critical importance. Applying both filters at the passage level could lead to the inclusion of extensive but irrelevant content, potentially diverting the attention of the LLM Reader in the subsequent Retrieval-Augmented Generation (RAG) system. For an in-depth discussion, see Appendix~\ref{appendix: case study}.

\section{Experiments}

\subsection{Datasets}
We evaluate LLM-AMT on three medical open-domain multiple-choice QA datasets as follows:

\paragraph{MedQA-USMLE and MedQA-MCMLE} \cite{jin2021disease} originate from professional medical board exams in the USA and Mainland China, where doctors are evaluated on their professional knowledge and ability to make clinical decisions.
% Questions in these exams are varied and generally require a deep understanding of related medical concepts from medical textbooks to answer.
In addition to the questions and corresponding answers, the datasets also provide associated medical textbook materials.
For the USMLE, the MedQA-USMLE dataset includes text extracted from a total of $18$ English medical textbooks used by USMLE candidates. For the MCMLE, the MedQA-MCMLE dataset features materials from $33$ simplified Chinese medical textbooks.
These are designated as the official textbooks for preparing for the medical licensing exam in Mainland China.

\paragraph{MedMCQA} \cite{pal2022medmcqa} encompasses a broad spectrum of 2,400 healthcare topics and 21 distinct medical subjects.
The diversity of questions contained within MedMCQA illustrates the challenges that are unique to this dataset.
As the questions are derived from both real-world scenarios and simulated examinations, they are meticulously crafted by human experts in the field. 
Consequently, these questions could serve as a comprehensive evaluation of a medical practitioner's professional competencies and expertise.

\begin{table}[t]
  \centering
  \scalebox{0.85}{
  \begin{tabular}{l >{\raggedleft\arraybackslash}p{1.3cm} >{\raggedleft\arraybackslash}p{1.3cm} >{\raggedleft\arraybackslash}p{1.3cm}}
    \toprule
    Question \# & \textbf{MedQA-USMLE} & \textbf{MedQA-MCMLE} & \textbf{Med-MCQA} \\
    % Question \# & MedQA-USMLE & MedQA-MCMLE & MedMCQA \\
    \midrule
    Train & 10,178 & 27,400 & 182,822 \\
    Dev   & 1,272  & 3,425  & 4,183 \\
    Test  & 1,273  & 3,426  & 6,150 \\
    \bottomrule
  \end{tabular}}
  \caption{Number of Questions in MedQA-USMLE, MedQA-MCMLE, and MedMCQA}
  \vspace{-2ex}
  \label{tab:datasets_summary}
\end{table}

Table~\ref{tab:datasets_summary} shows the detail of train/dev/test splits of the datasets.
We evaluate our pipeline and conduct ablation studies on the test sets of each dataset.

\subsection{Baselines}

Our evaluations encompass two primary categories of models. The first group consists of the \textit{Closed-Book Models}, which are pre-trained or fine-tuned specifically for the medical domain. These models rely on their internal knowledge and do not access external databases or texts during the question-answering process. Notable models in this category include \textbf{BioBERT}, \textbf{SciBERT}, \textbf{BioLinkBERT}, \textbf{PubmedBERT}, \textbf{Flan-PaLM (540B)}, \textbf{Meditron-70B}, \textbf{Med PaLM 2} \cite{lee2020biobert, beltagy2019scibert, yasunaga2022linkbert, gu2021domain, singhal2022large, chen2023meditron, singhal2023towards}. It is important to note that data marked with an asterisk\textsuperscript{*} were obtained directly from the respective authors' published works.

The second group, \textit{Wikipedia-Augmented Models}, leverages the knowledge embedded in Wikipedia to assist in the medical QA task. Key models in this category are \textbf{Variational ODQA} \cite{lievin2023variational}, \textbf{Codex 5-shot CoT} \cite{lievin2022can}. We have separately employed  \textbf{LLaMA-2-13B}, \textbf{GPT-3.5-Turbo}\footnote{In this paper, ``GPT-3.5'' denotes \textit{gpt-3.5-turbo-0125}. Similarly, references to ``GPT-4'' imply \textit{gpt-4-0125-preview.}.} and \textbf{GPT-4-Turbo} as readers, enhanced by the knowledge retrieved from Wikipedia to answer questions.

\subsection{Implementation Details}

We employ OpenAI’s GPT-3.5-Turbo as our LLM readers in different experiments. LLaMA-2-13B and GPT-4-Turbo are only used in the main result experiments of Table~\ref{tab:main_result}. Subsequent Ablation Studies only utilize GPT-3.5 as the generator. GPT-3.5-Turbo, accessed via its API\footnote{\url{https://platform.openai.com/docs/guides/gpt}}, handled query rewriting during the augmentation phase. In the evidence retrieval stage, SPLADE acts as our sparse retriever, DPR is the dense retriever, and we incorporate a cross-encoder for reranking. The MS-MARCO dataset \cite{nguyen2016ms} is our primary training source for our zero-shot model. Specifics related to fine-tuning, such as batch size, learning rate, and training rounds, can be found in the supplementary material.

\begin{table}[t]
  \centering
  \scalebox{0.8}{
  \begin{tabular}{l >{\centering\arraybackslash}p{1.3cm} >{\centering\arraybackslash}p{1.3cm} >{\centering\arraybackslash}p{1.9cm}}
    \toprule
    \textbf{Method} & \textbf{Retriever} & \textbf{MedQA-USMLE} & \textbf{MedMCQA} \\ % Removed line break for MedMCQA
    \midrule
    \multicolumn{4}{c}{Closed-Book Model} \\ 
    \midrule
    Random  & - & 20.0 & 25.0 \\
    BioBERT\textsuperscript{*} & - & 36.7 & 37.0 \\
    SciBERT\textsuperscript{*} & - & - & 39.0 \\
    BioLinkBERT\textsuperscript{*} & - & 45.1 & - \\
    PubmedBERT\textsuperscript{*} & - & 50.3 & 41.0 \\
    LLaMA    & - & 31.4 & 35.7 \\
    GPT-3.5  & - & 51.3 & 53.9 \\
    Flan-PaLM (540B)\textsuperscript{*} & - & 67.6 & - \\
    Meditron-70B\textsuperscript{*} & - & 70.2 & - \\
    GPT-4 & - & 81.7 & 70.5 \\
    Med-PaLM 2\textsuperscript{*} & - & 85.4 & 72.3 \\

    \midrule
    \multicolumn{4}{c}{Wikipedia-Augmented Model} \\ 
    \midrule
    Variational ODQA\textsuperscript{*}        & BM25+DPR & 55.0 & 62.9 \\
    Codex 5-shot CoT\textsuperscript{*}        & BM25 & 60.2 & 62.7 \\
    LLaMA + Wikipedia       & DPR & 38.6  & 40.5    \\
    LLaMA + Wikipedia       & HybTextR & 39.9  & 41.3   \\
    GPT-3.5 + Wikipedia     & DPR & 52.8 & 56.8 \\
    GPT-3.5 + Wikipedia     & HybTextR & 54.2 & 57.7 \\
    GPT-4 + Wikipedia       & DPR & 80.6 & 69.8 \\
    GPT-4 + Wikipedia       & HybTextR & 81.5 & 71.2 \\
    \midrule
    \multicolumn{4}{c}{Textbook-Augmented Model} \\ 
    \midrule
    LLM-AMT (LLaMA)      & HybTextR & 42.2 & 43.8 \\
    LLM-AMT (GPT-3.5)      & HybTextR & 67.9 & 65.5 \\
    LLM-AMT (GPT-4)      & HybTextR & \textbf{88.1} & \textbf{74.6} \\
    \bottomrule
  \end{tabular}
  }
  \caption{Performance of various state-of-the-art models on MedQA and MedMCQA datasets.}
  \label{tab:main_result}
\end{table}

\subsection{Main Result}

In Table~\ref{tab:main_result}, we compare various state-of-the-art models with our proposed pipeline on MedQA and MedMCQA datasets. 

Our experiments reveal that incorporating textbook knowledge with our proposed method significantly enhances the performance of GPT-3.5-Turbo and GPT-4-Turbo when compared to closed-book models. While Wikipedia is a rich information source, its content may be too generalized and often lacks the necessary depth for specialized fields such as medicine. Therefore, the smaller performance gains observed when utilizing Wikipedia as an external knowledge base may be due to the fact that these large language models have already incorporated Wikipedia data during pre-training. To explore the effectiveness of Wikipedia as a retrieval corpus, we employed two distinct retrievers: a publicly available pre-finetuned DPR from other researchers \cite{karpukhin2020dense}, and our own fine-tuned HybTextR system using the Wikipedia corpus as training data. Both methods indicated that a textbook corpus is more useful compared to Wikipedia for enhancing medical QA performance. This is evidenced by a 13.7\% increase over the \textit{GPT-3.5 + Wiki} for MedQA and a 7.8\% increase for MedMCQA, highlighting the significance of integrating deep, specialized medical knowledge overbroad, surface-level information sources.

Moreover, when leveraging the more sophisticated GPT-4 as the base model, our approach surpasses the performance of specialized closed-book models such as Flan-PaLM (540B) and Med-PaLM 2. This showcases the potential of combining large language models with targeted domain expertise, emphasizing the value of domain-specific knowledge in retrieval-augmented generation methods.

\subsubsection{Component Impact Analysis}
Our investigation into the LLM-AMT pipeline reveals the integral roles of the Textbook Retriever, Query Augmenter, and Knowledge Self-Refiner. In Table ~\ref{tab: incremental improvements}, we provide a unified analysis, demonstrating their collective impact on enhancing the model's performance on medical QA tasks, as evidenced in the MedQA-USMLE dataset and corroborated by similar trends in other datasets.

\begin{table}[ht]
\centering
\small
{
\begin{tabular}{l >{\centering\arraybackslash}p{1.1cm} >{\centering\arraybackslash}p{1.1cm} >{\centering\arraybackslash}p{1.1cm}}
\toprule
% \makecell[l]{\textbf{Method}} & \makecell[c]{\textbf{MedQA-}\\ \textbf{USMLE}} & \makecell[c]{\textbf{MedQA-}\\ \textbf{MCMLE}} & \makecell[c]{\textbf{MedMCQA}} \\
\textbf{Method} & \textbf{MedQA-USMLE} & \textbf{MedQA-MCMLE} & \textbf{Med-MCQA} \\
\midrule
GPT-3.5-Turbo & 51.3 & 58.2 & 53.9 \\
\midrule
+ retriever & 58.6 & 61.2 & 57.1 \\
\midrule
\makecell[l]{+ retriever \\ + augmented query} & 62.0 & 65.4 & 63.1 \\
\midrule
\makecell[l]{+ retriever \\ + knowledge self-refiner} & 63.9 & 68.1 & 64.4 \\
\midrule
\makecell[l]{+ retriever \\ + augmented query \\ + knowledge self-refiner} & 65.0 & 68.8 & 65.1 \\
\midrule
+ finetuned retriever & 61.2 & 62.3 & 58.7 \\
\midrule
\makecell[l]{+ finetuned retriever \\ + augmented query} & 64.1 & 68.9 & 63.4 \\
\midrule
\makecell[l]{+ finetuned retriever \\ + knowledge self-refiner} & 65.7 & 70.3 & 64.8 \\
\midrule
\makecell[l]{+ finetuned retriever \\ + augmented query \\ + knowledge self-refiner} & \textbf{67.9} & \textbf{72.6} & \textbf{65.5} \\
\bottomrule
\end{tabular}
}
\caption{Performance comparison (\% accuracy) of various approaches on three medical QA datasets. The table showcases the incremental improvements gained by integrating different components. Specifically, the retriever employed is HybTextR, and the LLM Reader is GPT-3.5-Turbo.}
\label{tab: incremental improvements}
\end{table}

\begin{enumerate}
    \item \textbf{Textbook Retriever (HybTextR)} serves as the cornerstone, providing a 7.3\% boost in accuracy by tapping into specialized medical literature for relevant information.

    \item \textbf{Query Augmenter} elevates recall by translating general inquiries into precise medical terminology and through query expansion to enhance relevant knowledge association, leading to a 3.4\% incremental accuracy gain. It ensures that the breadth of the search captures a wide spectrum of relevant evidence.

    \item \textbf{Knowledge Self-Refiner} complements by scrutinizing the relevance and usefulness of the retrieved information, fine-tuning precision, and contributing a further 1.9\% accuracy increase. It filters the evidence, sharpening the focus on the most pertinent medical facts.

    \item \textbf{Synergistic Effect}: The Query Augmenter and Knowledge Self-Refiner synergize to elevate LLM performance. The augmenter boosts knowledge recall, while the refiner improves precision, providing the LLM with high-quality, external medical knowledge. This synergy is crucial for handling complex medical queries. See the appendix \ref{appendix: case study} for a detailed case study on their interaction.
\end{enumerate}

% This integrated approach underscores the importance of both breadth and depth in knowledge retrieval and refinement, yielding a composite model that is well-equipped to address the complexities of medical question answering with superior accuracy.

\subsection{Ablation Study}

Here, we perform ablation studies on the query augmentation, the retrieval mechanisms, and the knowledge self-refinement strategy to refine and identify the most optimal configuration specifically tailored for question-answering tasks within the medical domain.

\subsubsection{Query Augmenter Components}
In this part, we performed a series of ablation experiments on the MedQA dataset to evaluate the efficacy of various components within our Query Augmenter framework. The detailed results are presented in Table~\ref{tab:query_augmenter_ablation}. Our findings demonstrate that each component contributes to the overall performance incrementally. These results underscore the synergistic effect of these components in improving the model's ability to understand and process complex medical queries.

\begin{table}[t]
\centering
\scalebox{0.85}{
\begin{tabular}{@{}l@{\hspace{20pt}}c@{}}
\toprule
\textbf{Method} & \textbf{Accuracy} \\
\midrule
GPT-3.5-Turbo & 51.3 \\
\midrule
\hspace{5pt}+ retriever & 58.6 \\
\hspace{5pt}+ query rewriting & 61.2\\
\hspace{5pt}+ query expansion & 62.0\\
\bottomrule
\end{tabular}}
\caption{Query Augmenter's ablation study}
\label{tab:query_augmenter_ablation}
\end{table}

\begin{table}[t]
    \centering
    \small
    \scalebox{0.9}{
    \setlength{\tabcolsep}{3pt}
    \begin{tabular}{lrr}
    \toprule
     & \multicolumn{1}{c}{\textbf{Zero-shot}} & \multicolumn{1}{c}{\textbf{Fine-tuned}} \\
    \cmidrule(r){2-2} \cmidrule(l){3-3}
     & \textbf{MedQA-USMLE} & \textbf{MedQA-USMLE} \\
    \midrule
    BM25 & 55.6 & -- \\
    Sparse & 57.4 & 59.3 \\
    Dense & 59.7 & 60.9 \\
    ColBERT & 58.2 & 61.5 \\
    Sparse + Dense & 60.1 & 62.7 \\
    Sparse + Rerank & 59.5 & 61.3 \\
    Dense + Rerank & 60.6 & 63.7 \\
    \textbf{HybTextR} & \textbf{62.0} & \textbf{64.1} \\
    \bottomrule
    \end{tabular}}
    \caption{Evaluation of Retrieval and Reranking Strategies on the Performance of LLM-AMT}
    \label{tab:retrievers_comparison}
\end{table}

\subsubsection{Textbook Retrievers}

In Table~\ref{tab:retrievers_comparison}, we evaluate the impact of different retrieval methods in our pipeline. The late-interaction ColBERT retriever notably achieves 58.2\% accuracy on MedQA-USMLE in a zero-shot scenario, surpassing standalone sparse and dense retrievers. A hybrid approach, combining dense and sparse retrievers, yields a higher accuracy of 60.1\%. 

Adding a reranker, the Dense + Rerank setup increases accuracy to 60.6\%. The HybTextR model, incorporating sparse, dense, and reranking, reaches the peak accuracy of 62.0\% on MedQA-USMLE, demonstrating the advantage of a layered retrieval approach in medical contexts. For similar experiments on MedQA-MCMLE and MedMCQA, refer to Appendix~\ref{appendix:exclusive_performance_medqa_mcmle_medmcqa}.

\subsubsection{Knowledge Self-Refiner Components}

\begin{table}[ht]
  \centering
  \scalebox{0.85}{
  \begin{tabular}{l c}
    \toprule
    \textbf{Configuration} & \textbf{Accuracy (\%)} \\
    \midrule
    w/o KSR & 64.1 \\
    + Relevance Filter & 65.8 \\
    + Usefulness Filter & 66.2 \\
    + Full System & 67.9 \\
    \bottomrule
  \end{tabular}}
  \caption{Impact of KSR components on GPT-3.5-Turbo LLM Reader accuracy.}
  \label{tab:ksr_ablation}
\end{table}

In this part, the impact of Knowledge Self-Refiner (KSR) components on a GPT-3.5-Turbo LLM Reader is examined within the MedQA domain, supplemented by HybTextR and Query Augmentation strategies. As illustrated in Table~\ref{tab:ksr_ablation}, the Relevance Filter marginally increases accuracy, underscoring its role in identifying pertinent content. The Usefulness Filter, contributing a slightly higher gain, is instrumental in isolating content of practical value for responses. The concurrent application of both filters results in the highest accuracy, signifying the importance of multi-dimensional content refinement in medical question-answering.

\subsection{Further Discussion}
% Here we further discuss on \wenhu{please have an intro sentence to summarize}...

In this section, we discuss and further assess our models, particularly their performance in non-multiple-choice medical QA tasks.

% \subsubsection{Non-multiple-choice QA Task}

\begin{table}[ht]
\centering
\scalebox{0.85}{
\begin{tabular}{lcc}
\toprule
Tiers              & \textbf{GPT-3.5} & \textbf{LLM-AMT} \\
\midrule
Correct            & 27           & 36      \\
Mostly Correct     & 10           & 12       \\
Partially Correct  & 14           & 19      \\
Wrong              & 49           & 33      \\
\bottomrule
\end{tabular}}
\caption{Evaluation of the Non-multiple-choice Medical Question Answering Task. GPT-3.5 as the LLM Reader.}
\label{tab:non-multiple-choice}
\end{table}

To test medical QA models in a realistic scenario, we chose 100 varied questions from the MedQA-USMLE dataset and produced answers without seeing the options. Medical professionals evaluated the answer quality, ranking them as:
\begin{itemize}
    \item \textbf{Correct}: Accurate and complete.
    \item \textbf{Mostly Correct}: Generally accurate, with some details missing.
    \item \textbf{Partially Correct}: Contains correct aspects but lacks key information.
    \item \textbf{Wrong}: Inaccurate or irrelevant.
\end{itemize}

Our LLM-AMT model surpassed the GPT-3.5-Turbo baseline in the non-multiple-choice QA task, delivering 36 correct answers to the baseline's 27. Notably, LLM-AMT provided more partially correct answers (19 vs. 14) and fewer errors (33 vs. 49). This underscores the model's enhanced accuracy in the medical QA domain, as detailed in Table~\ref{tab:non-multiple-choice}. The superior performance of LLM-AMT in the non-multiple-choice QA task not only illustrates its advanced capabilities but also emphasizes its potential for practical application in real-world medical scenarios. Such advancements can be instrumental in aiding medical professionals with more accurate and reliable information.

\section{Related Work}
In this section, we provide an overview of the related work in biomedical QA, retrieval-augmented QA, and text retrieval.
\subsection{Biomedical question answering}

Biomedical QA plays a pivotal role in clinical decision support \cite{ely2005answering} and the acquisition of biomedical knowledge \cite{jin2022biomedical}. With the rise of pre-trained language models (LMs), there's been a significant uptick in performance and the emergence of new capabilities across various natural language processing (NLP) tasks \cite{chowdhery2022palm, chung2022scaling, wei2022chain, wei2022emergent}. Nevertheless, these auto-regressive LLMs, when applied in domains like medicine and healthcare that require intensive knowledge or reasoning, are prone to generating hallucinations and erroneous content. Combining external knowledge sources with LLMs is a promising approach to counteract these pitfalls \cite{mialon2023augmented}.

\subsection{Retrieval Augmented Generation}

The retrieval-augmented generation paradigm, originating from the DrQA framework by \citeauthor{chen2017reading}, initially used heuristic retrievers like TF-IDF to source evidence from Wikipedia, followed by a neural model to extract answers. This methodology was advanced by DPR \cite{karpukhin2020dense}, using pre-trained transformers like BERT for retrieval and reading. Retrieval Augmented Generation (RAG) \cite{lewis2020retrieval} further evolved the approach by shifting from answer extraction to generation, enabling free-form text creation. Advances in RAG have explored retrieval as a critical tool for augmentation, with \citeauthor{schick2023toolformer}, \citeauthor{luo2023sail}, and \citeauthor{asai2023self} targeting enhanced information sourcing mechanisms. Moreover, \citeauthor{yan2024corrective} explored and designed corrective strategies for RAG to bolster generation robustness. Concurrently, models like REALM \cite{guu2020retrieval} and RETRO \cite{borgeaud2022improving} integrated retrieval during the pre-training phase. Recently, Large Language Models (LLMs) have been incorporated into this framework, as seen in REPLUG \cite{shi2023replug} and IC-RALM \cite{ram2023context}. While prior work on RAG primarily addressed general knowledge, this study introduces the first application of RAG to medical literature, harnessing a vast collection of medical textbooks. Our innovative knowledge self-refinement strategies enhance the fidelity of retrieved information, marking the first refinement of RAG's retrieval component for elevated performance in the medical domain.

\subsection{Neural Text Retrieval}

Recent progress in Neural Retrieval with bi-encoder architectures surpasses traditional methods like BM25/TF-IDF. This technique encodes queries and documents independently using pre-trained transformers, measuring similarity with embedding distances.
Neural retrieval can be categorized into \textit{dense retrieval} (e.g., DPR~\cite{karpukhin2020dense}, ANCE~\cite{xiong2020approximate}, CoCondenser~\cite{gao2021condenser}), \textit{sparse retrieval} (e.g., DeepImpact~\cite{mallia2021learning}, uniCOIL~\cite{lin2021few}, SPLADE~\cite{formal2021splade}), and \textit{late interaction retrieval} (e.g., ColBERT~\cite{khattab2020colbert}, COIL~\cite{gao2021coil}), based on the type of embedding used. 
In this study, we apply these neural retrieval methods to medical textbook retrieval, assessing their domain-specific effectiveness beyond standard corpora.

\section{Conclusion}

We introduced LLM-AMT, a novel pipeline optimized for medical tasks, harnessing authoritative medical textbooks to enhance LLMs' accuracy and professionalism. Empirical evaluations reinforced the value of integrating domain-specific textbooks with LLMs, providing an avenue for future studies. Further, our ablation study delineated the significance of external knowledge retrieval, query augmentation, and knowledge self-refinement strategy within our proposed architecture. These findings set a precedent for advancing specialized domain-aware models, especially in the context of medical informatics and healthcare AI applications.

\section{Limitations}

\subsection{Model Explainability}
One significant limitation of the LLM-AMT system lies in its inherent lack of explainability. While the integration of authoritative medical textbooks enhances the model's responses, the reasoning behind these responses often remains opaque. Medical decision-making demands a high degree of transparency; however, as with many large language models, the LLM-AMT operates as a ``black box''. This poses a challenge in clinical settings, where explanations for diagnoses or treatment recommendations are crucial for trust and accountability. Incorrect or unexplained advice from the model could lead to misdiagnosis or inappropriate treatment, endangering patient health and potentially eroding trust in AI-assisted medical systems. The model's inability to provide detailed explanations for its conclusions can be a significant barrier to its adoption in practice.

\subsection{Interactive Question-Answering}
Another limitation is the system's capacity for interactive QA. In real-world medical practice, diagnostic and treatment processes involve nuanced communications with patients, requiring a deep understanding of individual circumstances, empathetic engagement, and the ability to ask follow-up questions for clarification. The LLM-AMT, despite its advancements, cannot fully replicate this level of interaction. The model might not adequately handle the subtleties of patient-specific narratives or the dynamic nature of medical conversations. Thus, while LLM-AMT can provide informative responses, its interactive capabilities are limited in comparison to the rich, two-way communication typically found in patient-clinician interactions.

\bibliography{custom}

\appendix

\section{Appendix}
\label{sec:appendix}

\subsection{Exclusive Performance Evaluation on MedQA-MCMLE and MedMCQA}
\label{appendix:exclusive_performance_medqa_mcmle_medmcqa}

Table~\ref{tab:performance_medqa_mcmle_medmcqa} presents the performance of various retrieval and reranking strategies exclusively on the MedQA-MCMLE and MedMCQA datasets. These datasets pose distinct challenges compared to MedQA-USMLE and thus merit a separate analysis. The results provide insights into the generalizability and robustness of the methods when confronted with different types of medical question-answering datasets. The HybTextR method, in particular, shows a consistently strong performance, suggesting its potential as a versatile tool for medical information retrieval tasks.

\subsection{Fine-tuning Hyperparameters for Retrievers}

Table \ref{tab: hyperparameters} presents the hyperparameters used for fine-tuning retrievers. In the \textit{Seq Length} column of Table \ref{tab: hyperparameters}, the notation \(32 + 220\) for the ColBERT model indicates that the maximum length for the query is set to 32, while the length for the passage is 220. For the Reranker model, the input sequence is structured as \texttt{[CLS]} token followed by the query, then a \texttt{[SEP]} token, and finally the passage. Therefore, its sequence length is calculated as \(1 + 126 + 1 + 384\), which sums up to 512.

\begin{table}[ht]
\centering
\setlength{\tabcolsep}{4pt} % Adjust the padding between table columns
\begin{tabular}{lccc}
\toprule
\textbf{Model} & \textbf{Batch} & \textbf{Seq Len} & \textbf{LR} \\
\midrule
Splade & 64 & 256 & $2 \times 10^{-5}$ \\
DPR & 8 & 256 & $1 \times 10^{-5}$ \\
ColBERT & 32 & 32+220 & $3 \times 10^{-6}$ \\
Reranker & 8 & 126+384 & $8 \times 10^{-6}$ \\
\bottomrule
\end{tabular}
\caption{Hyperparameters for fine-tuning.}
\label{tab: hyperparameters}
\end{table}

\begin{table*}[ht]
    \centering
    \small
    \setlength{\tabcolsep}{3pt}
    \begin{tabular}{lrrrr}
    \toprule
     & \multicolumn{2}{c}{\textbf{Zero-shot}} & \multicolumn{2}{c}{\textbf{Fine-tuned}} \\
    \cmidrule(r){2-3} \cmidrule(l){4-5}
     & \textbf{MedQA-MCMLE} & \textbf{MedMCQA} & \textbf{MedQA-MCMLE} & \textbf{MedMCQA} \\
    \midrule
    BM25 & 59.7 & 55.2 & -- & -- \\
    Sparse & 60.4 & 57.5 & 62.9 & 59.6 \\
    Dense & 61.0 & 57.7 & 63.8 & 59.3 \\
    ColBERT & 62.4 & 58.1 & 64.1 & 60.4 \\
    Sparse + Dense & 64.9 & 58.7 & 65.5 & 61.9 \\
    Sparse + Rerank & 63.8 & 59.2 & 65.2 & 62.8 \\
    Dense + Rerank & \textbf{65.4} & 61.8 & 65.3 & 64.6 \\
    \textbf{HybTextR} & 64.4 & \textbf{63.1} & \textbf{68.9} & \textbf{65.2} \\
    \bottomrule
    \end{tabular}
    \caption{Evaluation of Retrieval and Reranking Strategies on MedQA-MCMLE and MedMCQA Datasets}
    \label{tab:performance_medqa_mcmle_medmcqa}
\end{table*}

\subsection{Models}
We use the following model:
\begin{itemize}
    \item \textbf{DPR}, which uses BERT-base as the backbone and has 110M parameters. It is under the CC-BY-NC 4.0 License.
    \item \textbf{SPLADE}, which uses BERT-base as the backbone and has 110M parameters. It is under the CC BY-NC-SA 4.0. License.
    \item \textbf{ColBERT}, which uses BERT-base as the backbone and has 110M parameters. It is under the MIT License.
    \item \textbf{LLaMA-2-13B}, 13B parameters, under the Llama 2 Community License Agreement.
    \item \textbf{GPT-3.5-Turbo and GPT-4-Turbo}, which are not open-source and can only be accessed via API requests.
\end{itemize}

\subsection{Datasets}
We use the following datasets:
\begin{itemize}
    \item \textbf{MedQA}, which is under the MIT Licenses. The intended purpose of the MedQA dataset is to support and advance research in the area of natural language processing (NLP) and information retrieval (IR) within the medical domain. MedQA is composed of both English and Chinese questions and answers. While the dataset predominantly features clinical scenarios and medical knowledge representations, demographic information of the represented groups is not explicitly detailed due to the nature of the data.
    \item \textbf{MedMCQA}, which is under the MIT License for non-commercial research purposes. The MedMCQA dataset spans a broad range of medical domains, including but not limited to cardiology, oncology, pediatrics, neurology, and infectious diseases. Each domain is represented with questions and answers that reflect the diversity of medical knowledge. The dataset is primarily in English, ensuring that the findings of our research are directly applicable to English-language medical question-answering systems.
\end{itemize}

\subsection{AI Assistance in Writing}
In the preparation of this manuscript, we utilized an AI language model, specifically ChatGPT, to assist with grammar checking and refining the expressions used in our writing. This utilization was confined to ensuring linguistic accuracy and enhancing readability, without influencing the scientific content or the originality of the research findings presented. The contribution of ChatGPT was strictly as a supportive tool for language polishing, and all final decisions regarding the manuscript content were made by the human authors.

\subsection{Full List of Instructions For GPT-3.5-Turbo}
\label{appendix:ins_list}
In Table~\ref{tab:ins_list_table}, we list the instructions we used in LLM-AMT.

\begin{table*}[htbp]
\centering
\scalebox{0.95}{
\begin{tabular}{p{\linewidth}}
\toprule
\textbf{Query Rewriting:} \\
Question: XXX \\
Please reformulate the given question by employing precise medical terminology. Focus on capturing the essence of the patient's symptoms and conditions in a generalized form that reflects common clinical descriptions. Avoid using colloquial language and ensure that the rewritten query is clear, concise, and can be universally understood in a professional medical context. \\
\midrule 
\textbf{Query Expansion:} \\
Question: XXX \\
Assume the role of a medical doctor and expand upon the initial query. Conduct a systematic analysis by dissecting the question into its medical components. Then, elaborate on each component with detailed medical insights that collectively build a comprehensive understanding of the underlying health issue. Proceed methodically to ensure that each step of your explanation contributes to a logically structured answer. \\
\midrule 
\textbf{Relevance Filter:} \\
Retrieved Passage: XXX \\
x: XXX \\
Examine the retrieved passages above carefully. Determine if each passage pertains to the context of the specific query represented by 'x'. Respond with 'Yes' if a passage is relevant and contributes meaningful information to the query, or 'No' if it does not relate to the query or provide valuable insight. Please answer with 'Yes' or 'No' only for each passage assessed.\\
\midrule 
\textbf{Usefulness Filter:} \\
Retrieved Information: XXX \\
x: XXX \\
Review the information retrieved above and evaluate its utility in addressing the question represented by 'x'. Provide a response of 'Yes' if the information is pertinent and aids in formulating a comprehensive answer, or 'No' if it lacks relevance or does not contribute to a substantive response to the question. Respond with a singular 'Yes' or 'No' for the usefulness of each piece of information.\\
\midrule
\textbf{LLM Reader Instruction:} \\
Medical Knowledge: XXX \\
Question: XXX \\
Using the medical knowledge provided, please answer the following medical question with a chain-of-thought approach. Break down your reasoning into clear, logical steps that detail your clinical thought process from initial hypothesis formation through to the final conclusion, similar to how a medical professional would approach a diagnostic challenge. Your answer should not only be informed by the medical knowledge but also transparent in the reasoning that led to your conclusion. \\
\bottomrule
\end{tabular}}
\caption{Full List of Instructions For GPT-3.5-Turbo}
\label{tab:ins_list_table}
\end{table*}

\subsection{Case Study}
\label{appendix: case study}

As part of our comprehensive case study, Tables~\ref{tab:case_study_table}, \ref{tab:case_study_table_2}, \ref{tab:case_study_table_3}, and \ref{tab:case_study_table_4} present detailed input and output data corresponding to each component depicted in the overview (Figure~\ref{overview}). These tables include the full question and options, retrieved passages from the textbook, retrieved passages from Wikipedia, the rewritten query, the expanded query, the results from the knowledge self-refinement stage, and the final refined knowledge. This granular view provides clear insight into the information processing pipeline and the effectiveness of each module.
\newpage

\begin{table*}[ht]
\centering
\begin{tabularx}{\textwidth}{X}
\toprule
\textbf{Question:} \\
Two weeks after undergoing an emergency cardiac catheterization with stenting for unstable angina pectoris, a 61-year-old man presents with decreased urinary output and malaise. He has type 2 diabetes mellitus and osteoarthritis of the hips. His medications prior to admission were insulin and naproxen, and he was started on aspirin, clopidogrel, and metoprolol after the coronary intervention. His current vitals are: temperature 38°C (100.4°F), pulse 93/min, blood pressure 125/85 mm Hg. Examination reveals mottled, reticulated purplish discoloration of the feet. Laboratory studies show: Hemoglobin count 14 g/dL, Leukocyte count 16,400/mm3, Segmented neutrophils 56\%, Eosinophils 11\%, Lymphocytes 31\%, Monocytes 2\%, Platelet count 260,000/mm3, Erythrocyte sedimentation rate 68 mm/h, Serum Urea nitrogen 25 mg/dL, Creatinine 4.2 mg/dL. Renal biopsy shows intravascular spindle-shaped vacuoles. The question is: Which of the following is the most likely cause of this patient's symptoms? \\
\textbf{Options:} \\
A) Renal papillary necrosis \\
B) Cholesterol embolization \\
C) Eosinophilic granulomatosis with polyangiitis \\
D) Polyarteritis nodosa \\
\midrule 
\textbf{Ground Truth:} \\
Cholesterol embolization (Choice B) \\
\midrule
\textbf{Rewritten Query:} \\
What is the most likely etiology of the oliguria and malaise in this patient with a history of emergency cardiac catheterization, type 2 diabetes mellitus, and \underline{\textbf{osteoarthritis of the hips}}, along with \underline{\textbf{mottled}}, \underline{\textbf{reticulated purplish discoloration of the feet}}, \underline{\textbf{elevated creatinine levels}}, and \underline{\textbf{intravascular spindle-shaped vacuoles on renal biopsy}}? \\
\midrule 
\textbf{Expanded Query:} \\
Given the clinical scenario and lab findings, we should consider several factors to systematically approach this medical question. Let's analyze each option step by step.

A. Renal Papillary Necrosis: Presence of diabetes could predispose the patient to this condition, but the biopsy's intravascular spindle-shaped vacuoles are not characteristic of renal papillary necrosis, and the discoloration of the feet isn't a typical symptom.

B. Cholesterol Embolization: Recent cardiac catheterization could lead to this condition; however, one would need to consider other signs and symptoms and correlate them with further tests to rule out or confirm this diagnosis. The biopsy finding and the patient's presentation could be suggestive, but are not definitive without further evidence.

C. Eosinophilic Granulomatosis with Polyangiitis (EGPA): While the patient has eosinophilia, the absence of asthma or typical respiratory symptoms and granuloma formation on biopsy makes this option less likely.

D. Polyarteritis Nodosa (PAN): The absence of typical symptoms and biopsy findings associated with PAN, alongside the presence of eosinophilia, which isn't commonly seen in PAN, makes this diagnosis less likely.

In summary, while certain features may seem to align with some of the conditions, the evidence is not conclusive for any of the options based on the information presented. Further diagnostic workup would be essential to come to a definitive diagnosis. \\
\bottomrule
\caption{Case Study}
\label{tab:case_study_table}
\end{tabularx}
\end{table*}

\begin{table*}[ht]
\centering
\begin{tabularx}{\textwidth}{X}
\toprule
\textbf{Retrieved Passages from Wikipedia:} \\
Cardiac catheterization is a medical procedure used to diagnose and treat certain cardiovascular conditions. During cardiac catheterization, a long thin tube called a catheter is inserted in an artery or vein in your groin, neck, or arm and threaded through your blood vessels to your heart. Using this catheter, doctors can then perform diagnostic tests as part of a cardiac cath work-up. Procedures can include coronary angiography, as well as treatments like coronary angioplasty or stenting to open up narrowed or blocked arteries. {\color{red}\ding{55}}\\
Type 2 diabetes mellitus is a chronic metabolic disorder characterized by high blood sugar, insulin resistance, and relative lack of insulin. It is often associated with obesity and a sedentary lifestyle and can lead to severe health complications if not managed effectively. Management includes lifestyle changes, monitoring blood sugar levels, and medication or insulin therapy. {\color{red}\ding{55}}\\

Osteoarthritis of the hips is a degenerative joint disease that affects the hip joints. It occurs when the protective cartilage that cushions the tops of the bones wears down over time, which can lead to pain, swelling, and reduced range of motion in the hip. While the exact cause is unknown, factors such as age, joint injury, obesity, and genetics can increase the risk of developing hip osteoarthritis. {\color{red}\ding{55}}\\

Mottled, reticulated purplish discoloration of the feet is a skin condition that can indicate a variety of circulatory problems. The appearance of a lace-like pattern of purplish discoloration, known as livedo reticularis, may be due to the occlusion of small blood vessels or could be a reaction to cold temperatures. In certain cases, it can be associated with more serious underlying conditions, such as cholesterol embolization syndrome or vasculitis. {\color{red}\ding{55}}\\

Elevated creatinine levels in the blood can indicate impaired kidney function or kidney disease. Creatinine is a waste product produced by muscles from the breakdown of a compound called creatine. Normally, the kidneys filter out creatinine from the blood and excrete it in urine. High levels of creatinine may signal that the kidneys are not working properly. {\color{red}\ding{55}}\\

Intravascular spindle-shaped vacuoles observed on renal biopsy can be indicative of a pathological process affecting the kidneys. These structures could suggest the presence of intravascular material, such as cholesterol emboli, which can result from the dislodgement of atheromatous plaques following procedures like cardiac catheterization. The presence of such vacuoles requires further pathological examination to determine the precise cause and appropriate treatment. {\color{red}\ding{55}}\\
\textbf{...} \\
\midrule

\textbf{Retrieved Passages from Textbooks:}

Cholesterol embolization, also known as cholesterol crystal embolism (CCE) or atheroembolism, is a pathology that arises from the showering of cholesterol crystals from atherosclerotic plaques into the systemic circulation. The condition can result in a vasculopathy that is often systemic and can lead to end-organ damage due to {\color{green}the occlusion of small to medium-sized vessels}.
Pathophysiology. 
Cholesterol embolization occurs when cholesterol crystals dislodge from plaques in larger arteries, such as the aorta, and travel through the bloodstream to smaller blood vessels, where they become lodged and cause ischemia. The release of cholesterol crystals can occur spontaneously, but it is most commonly associated with iatrogenic causes such as {\color{green}cardiac catheterization}, vascular surgery, and anticoagulation therapy. 
Clinical Presentation. 
The clinical manifestations of cholesterol embolization are variable and depend on the organ systems involved. Common symptoms include:
Cutaneous: Livedo reticularis, {\color{green}cyanotic toes, or gangrene due to skin and soft tissue ischemia.}
Renal: Acute or chronic renal insufficiency presenting as {\color{green}an increase in serum creatinine, proteinuria, and hematuria}.
Gastrointestinal: Abdominal pain, gastrointestinal bleeding, and pancreatitis.
Neurological: Transient ischemic attacks, strokes, and multi-infarct dementia.  {\color{green}\ding{51}}\\

\bottomrule
\end{tabularx}
\caption{Continued Case Study}
\label{tab:case_study_table_2}
\end{table*}

\begin{table*}[ht]
\centering
\begin{tabularx}{\textwidth}{X}
\toprule
\textbf{Retrieved Passages from Textbook:} \\

Diagnosis of Cholesterol embolization is often challenging due to the nonspecific nature of the symptoms. Cholesterol embolization is frequently underdiagnosed or misdiagnosed. Laboratory findings may include {\color{green}eosinophilia and elevated erythrocyte sedimentation rate (ESR)}. Definitive diagnosis typically requires histological examination of affected tissue, which would reveal biconvex, needle-shaped clefts within the vessel lumen, indicative of cholesterol crystals. 
There is no specific treatment for cholesterol embolization. Management is primarily supportive and focuses on treating complications and symptomatic relief. Statin therapy may be used to stabilize atherosclerotic plaques and reduce the risk of further embolization. In cases related to medication, such as anticoagulants or thrombolytics, discontinuation or adjustment of the drugs may be required. 
The prognosis for patients with cholesterol embolization varies and can range from benign to severe, life-threatening multiorgan failure. Early recognition and management of the condition are crucial to improving outcomes. {\color{green}\ding{51}}\\

Cardiac Catheterization
Cardiac catheterization is an invasive diagnostic procedure that provides comprehensive information about the heart and surrounding blood vessels. This procedure involves the insertion of a catheter, which is a thin, flexible tube, into the coronary arteries and heart chambers under X-ray guidance.
Indications
Cardiac catheterization is typically indicated for the following purposes: Diagnosing coronary artery disease
Evaluating heart muscle function
Determining the need for further treatment (such as angioplasty or coronary artery bypass surgery)
Assessing the effectiveness of prior procedures
Measuring the pressure within the heart chambers
Diagnosing congenital heart abnormalities {\color{red}\ding{55}} \\

\textbf{...} \\

\midrule 
\textbf{Passages after Relevance Filter:} 

Cholesterol embolization, also known as cholesterol crystal embolism (CCE) or atheroembolism, is a pathology that arises from the showering of cholesterol crystals from atherosclerotic plaques into the systemic circulation. The condition can result in a vasculopathy that is often systemic and can lead to end-organ damage due to {\color{green}the occlusion of small to medium-sized vessels}.
Pathophysiology. 
Cholesterol embolization occurs when cholesterol crystals dislodge from plaques in larger arteries, such as the aorta, and travel through the bloodstream to smaller blood vessels, where they become lodged and cause ischemia. The release of cholesterol crystals can occur spontaneously, but it is most commonly associated with iatrogenic causes such as {\color{green}cardiac catheterization}, vascular surgery, and anticoagulation therapy. 
Clinical Presentation. 
The clinical manifestations of cholesterol embolization are variable and depend on the organ systems involved. Common symptoms include:
Cutaneous: Livedo reticularis, {\color{green}cyanotic toes, or gangrene due to skin and soft tissue ischemia.}
Renal: Acute or chronic renal insufficiency presenting as {\color{green}an increase in serum creatinine, proteinuria, and hematuria}.
Gastrointestinal: Abdominal pain, gastrointestinal bleeding, and pancreatitis.
Neurological: Transient ischemic attacks, strokes, and multi-infarct dementia. \\

Diagnosis of Cholesterol embolization is often challenging due to the nonspecific nature of the symptoms. Cholesterol embolization is frequently underdiagnosed or misdiagnosed. Laboratory findings may include {\color{green}eosinophilia and elevated erythrocyte sedimentation rate (ESR)}. Definitive diagnosis typically requires histological examination of affected tissue, which would reveal biconvex, needle-shaped clefts within the vessel lumen, indicative of cholesterol crystals. 
There is no specific treatment for cholesterol embolization. Management is primarily supportive and focuses on treating complications and symptomatic relief. Statin therapy may be used to stabilize atherosclerotic plaques and reduce the risk of further embolization. In cases related to medication, such as anticoagulants or thrombolytics, discontinuation or adjustment of the drugs may be required. 
The prognosis for patients with cholesterol embolization varies and can range from benign to severe, life-threatening multiorgan failure. Early recognition and management of the condition are crucial to improving outcomes. \\

\bottomrule
\end{tabularx}
\caption{Continued Case Study}
\label{tab:case_study_table_3}
\end{table*}

\begin{table*}[ht]
\centering
\begin{tabularx}{\textwidth}{X}
\toprule
\textbf{Segments after Usefulness Filter:} 

Cholesterol embolization, also known as cholesterol crystal embolism (CCE) or atheroembolism, is a pathology that arises from the showering of cholesterol crystals from atherosclerotic plaques into the systemic circulation. The condition can result in a vasculopathy that is often systemic and can lead to end-organ damage due to {\color{green}the occlusion of small to medium-sized vessels}. {\color{green}\ding{51}}\\

Pathophysiology. 
Cholesterol embolization occurs when cholesterol crystals dislodge from plaques in larger arteries, such as the aorta, and travel through the bloodstream to smaller blood vessels, where they become lodged and cause ischemia. The release of cholesterol crystals can occur spontaneously, but it is most commonly associated with iatrogenic causes such as {\color{green}cardiac catheterization}, vascular surgery, and anticoagulation therapy. {\color{green}\ding{51}}\\

Clinical Presentation. 
The clinical manifestations of cholesterol embolization are variable and depend on the organ systems involved. Common symptoms include:
Cutaneous: Livedo reticularis, {\color{green}cyanotic toes, or gangrene due to skin and soft tissue ischemia.}
Renal: Acute or chronic renal insufficiency presenting as {\color{green}an increase in serum creatinine, proteinuria, and hematuria}.
Gastrointestinal: Abdominal pain, gastrointestinal bleeding, and pancreatitis.
Neurological: Transient ischemic attacks, strokes, and multi-infarct dementia. {\color{green}\ding{51}}\\

Diagnosis of Cholesterol embolization is often challenging due to the nonspecific nature of the symptoms. Cholesterol embolization is frequently underdiagnosed or misdiagnosed. Laboratory findings may include {\color{green}eosinophilia and elevated erythrocyte sedimentation rate (ESR)}. Definitive diagnosis typically requires histological examination of affected tissue, which would reveal biconvex, needle-shaped clefts within the vessel lumen, indicative of cholesterol crystals. {\color{green}\ding{51}}\\

There is no specific treatment for cholesterol embolization. Management is primarily supportive and focuses on treating complications and symptomatic relief. Statin therapy may be used to stabilize atherosclerotic plaques and reduce the risk of further embolization. In cases related to medication, such as anticoagulants or thrombolytics, discontinuation or adjustment of the drugs may be required. {\color{red}\ding{55}} \\

The prognosis for patients with cholesterol embolization varies and can range from benign to severe, life-threatening multiorgan failure. Early recognition and management of the condition are crucial to improving outcomes. {\color{red}\ding{55}} \\

\midrule
\textbf{Refined Knowledge:}

Cholesterol embolization, also known as cholesterol crystal embolism (CCE) or atheroembolism, is a pathology that arises from the showering of cholesterol crystals from atherosclerotic plaques into the systemic circulation. The condition can result in a vasculopathy that is often systemic and can lead to end-organ damage due to {\color{green}the occlusion of small to medium-sized vessels}. 
Pathophysiology. 
Cholesterol embolization occurs when cholesterol crystals dislodge from plaques in larger arteries, such as the aorta, and travel through the bloodstream to smaller blood vessels, where they become lodged and cause ischemia. The release of cholesterol crystals can occur spontaneously, but it is most commonly associated with iatrogenic causes such as {\color{green}cardiac catheterization}, vascular surgery, and anticoagulation therapy.
Clinical Presentation. 
The clinical manifestations of cholesterol embolization are variable and depend on the organ systems involved. Common symptoms include:
Cutaneous: Livedo reticularis, {\color{green}cyanotic toes, or gangrene due to skin and soft tissue ischemia.}
Renal: Acute or chronic renal insufficiency presenting as {\color{green}an increase in serum creatinine, proteinuria, and hematuria}.
Gastrointestinal: Abdominal pain, gastrointestinal bleeding, and pancreatitis.
Neurological: Transient ischemic attacks, strokes, and multi-infarct dementia. Diagnosis of Cholesterol embolization is often challenging due to the nonspecific nature of the symptoms. Cholesterol embolization is frequently underdiagnosed or misdiagnosed. Laboratory findings may include {\color{green}eosinophilia and elevated erythrocyte sedimentation rate (ESR)}. Definitive diagnosis typically requires histological examination of affected tissue, which would reveal biconvex, needle-shaped clefts within the vessel lumen, indicative of cholesterol crystals. \\

\bottomrule
\end{tabularx}
\caption{Continued Case Study}
\label{tab:case_study_table_4}
\end{table*}

\end{document}